\documentclass{article}

\usepackage{PRIMEarxiv}

\usepackage[utf8]{inputenc} % allow utf-8 input
\usepackage[T1]{fontenc}    % use 8-bit T1 fonts
\usepackage{hyperref}       % hyperlinks
\usepackage{url}            % simple URL typesetting
\usepackage{booktabs}       % professional-quality tables
\usepackage{amsfonts}       % blackboard math symbols
\usepackage{nicefrac}       % compact symbols for 1/2, etc.
\usepackage{microtype}      % microtypography
\usepackage{lipsum}
\usepackage{fancyhdr}       % header
\usepackage{multirow}
\usepackage{xcolor}%
\usepackage{algorithm}%
\usepackage{algorithmicx}%
\usepackage{algpseudocode}%
\usepackage{graphicx}%

\title{Knowledge-Enhanced Facial Expression Recognition with Emotional-to-Neutral Transformation
}

\author{Hangyu Li$^1$, Yihan Xu$^2$, Jiangchao Yao$^{3,4}$, Nannan Wang$^{2*}$, Xinbo Gao$^5$, Bo Han$^1$\\
{$^1$}TMLR Group, Department of Computer Science, Hong Kong Baptist University, \\
Hong Kong SAR, China.\\
{$^2$}State Key Laboratory of Integrated Services Networks, Xidian University, \\
Xi’an, 710071, Shaanxi, China.\\
{$^3$}Cooperative Medianet Innovation Center, Shanghai Jiao Tong University, \\
Shanghai, 200240, China.\\
{$^4$}Shanghai AI Laboratory, Shanghai, 200232, China.\\
{$^5$}Chongqing Key Laboratory of Image Cognition, Chongqing University of Posts and Telecommunications, \\
Chongqing, 400065, China.\\
{\tt\small hangyuli.xidian@gmail.com; yihanxu@stu.xidian.edu.cn; Sunarker@sjtu.edu.cn;}\\
{\tt\small nnwang@xidian.edu.cn; gaoxb@cqupt.edu.cn; bhanml@comp.hkbu.edu.hk}
}

\begin{document}
\maketitle

\begin{abstract}
Existing facial expression recognition (FER) methods typically fine-tune a pre-trained visual encoder using discrete labels. However, this form of supervision limits to specify the emotional concept of different facial expressions. In this paper, we observe that the rich knowledge in text embeddings, generated by vision-language models, is a promising alternative for learning discriminative facial expression representations. Inspired by this, we propose a novel knowledge-enhanced FER method with an emotional-to-neutral transformation. Specifically, we formulate the FER problem as a process to match the similarity between a facial expression representation and text embeddings. Then, we transform the facial expression representation to a neutral representation by simulating the difference in text embeddings from textual facial expression to textual neutral. Finally, a self-contrast objective is introduced to pull the facial expression representation closer to the textual facial expression, while pushing it farther from the neutral representation. We conduct evaluation with diverse pre-trained visual encoders including ResNet-18 and Swin-T on four challenging facial expression datasets. Extensive experiments demonstrate that our method significantly outperforms state-of-the-art FER methods. The code will be publicly available.
\end{abstract}

% keywords can be removed

\section{Introduction}
\label{sec:introduction}

Facial expression is an important part of nonverbal communication \cite{zhang2018facial}. By analyzing facial images, we can obtain various types of emotions, including surprise, fear, disgust, happiness, sadness, and anger. Facial expression recognition (FER) \cite{li2022survey} has been a long-lasting research area in computer vision and has achieved promising performance, with the core being learning discriminative facial expression representations. 

Since labeling facial expressions is a time-costly process \cite{Li_2017_CVPR}, existing FER methods generally fine-tune a visual encoder (\emph{e.g.}, ResNet \cite{he2016deep} and Swin Transformer \cite{liu2021swin}) pre-trained on the large-scale face recognition dataset MS-Celeb-1M \cite{guo2016ms}, for learning facial expression representations with limited training data \cite{zhang2022learn, Xue_2021_ICCV}. Then, they train a classifier to map facial expression representations to confidence scores. However, almost all existing methods learn representations using discrete labels, ignoring the emotional concept of different facial expressions \cite{Yang_2022_CVPR}. For example, given the discrete labels of fear ``2'' and anger ``6'', the differences in the emotional concept are not characterized. Therefore, there is still a need to design an appropriate supervision signal for facial expression representation learning. 

\begin{figure}[t]
  \centering
    \includegraphics[width=0.8\linewidth]{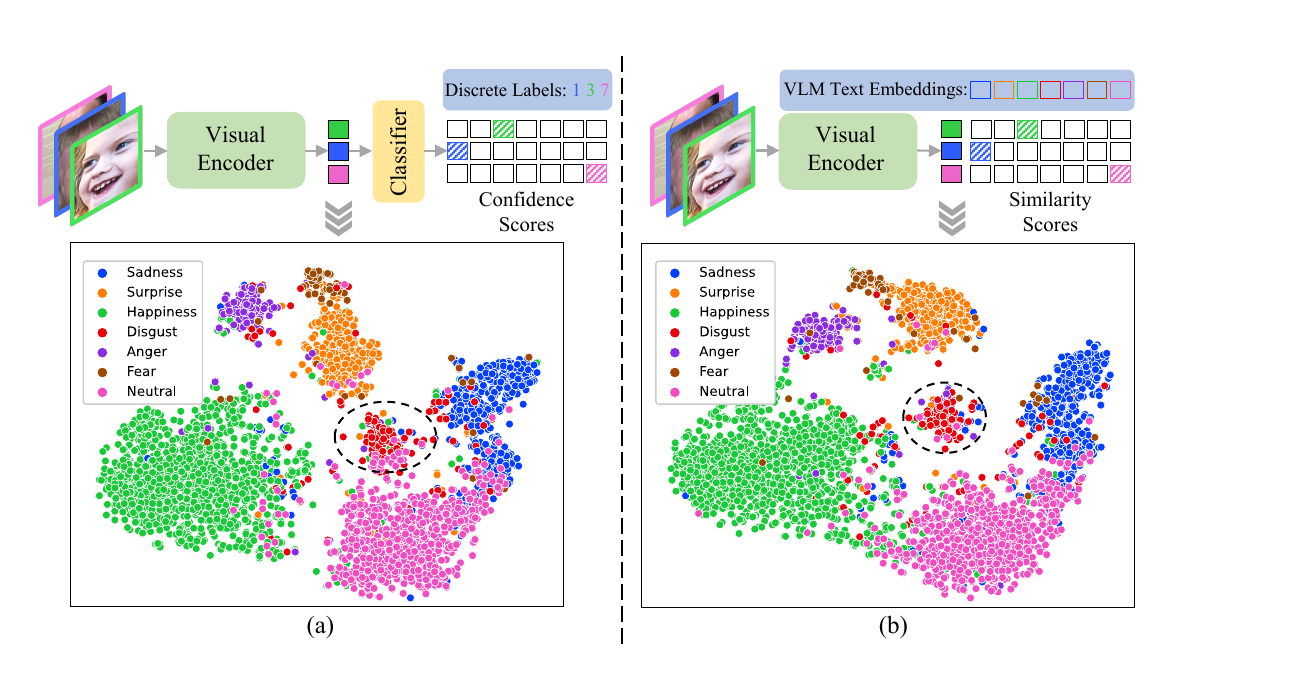}
   \caption{Illustration of facial expression recognition: (a) During fine-tuning ResNet-18 using discrete labels, a classifier is trained to map facial expression representations to confidence scores; (b) During fine-tuning ResNet-18 using VLM text embeddings, facial expression representations are compared with them for similarity scores. After fine-tuning, we use t-SNE \cite{maaten2008visualizing} to visualize the representation distribution of testing data from RAF-DB.}
   \label{fig:1}
\end{figure}

Recently, vision-language models (VLM) \cite{radford2021learning, jia2021scaling, li2022blip} have effectively learned visual concepts from their corresponding natural language. This is empowered by the alignment between visual representations and text embeddings, containing the rich knowledge. Inspired by this, we investigate whether VLM text embeddings as an external knowledge can better supervise facial expression representations. As shown in Figure~\ref{fig:1}, we observe that facial expression representations learned using text embeddings are more discriminative than those learned using discrete labels, \emph{e.g.}, the Disgust category. This observation implies that VLM text embeddings are more effective to guide facial expression representation learning.

In this work, we propose a knowledge-enhanced FER method with an emotional-to-neutral transformation. Specifically, we match the similarity between a facial expression representation and the text embeddings from the powerful VLM. Meanwhile, we draw inspiration from Russell’s Circumplex Model \cite{russell1980circumplex} to derive a neutral representation from the facial expression representation itself. To this end, we simulate the difference in text embeddings between textual facial expression and textual neutral. Then, we transform the facial expression representation to a neutral representation with this textual difference. Finally, we introduce a self-contrast objective to pull the facial expression representation closer to the textual facial expression and push it farther from the neutral representation. Briefly, it ensures that facial expression representations align more closely with the emotional knowledge in text embeddings. Overall, the main contributions of this work are summarized as follows:

\begin{itemize}
\item To the best of our knowledge, we first incorporate the rich knowledge in VLM text embeddings to fine-tune an arbitrary visual encoder for facial expression recognition.
\item We propose an emotional-to-neutral transformation along with a self-contrast objective to further enhance facial expression representations in a text-guided manner.
\item Extensive experiments on four challenging datasets show the effectiveness of VLM text embeddings for FER. In addition, our method achieves promising results compared to previous methods with diverse visual encoders.
\end{itemize}

The rest of this paper is organized as follows. Section~\ref{sec:related} gives the related work and the discussion between our work and existing methods. Then, we introduce the proposed method in Section~\ref{sec:method}. We further conduct experiments along with in-depth analysis in Section~\ref{sec:experiments}. Finally, the conclusion and the limitations in our work are given in Section~\ref{sec:conclusion}.

\section{Related Work}
\label{sec:related}

In this section, we briefly review facial expression recognition, VLM in facial expression recognition, and disentangled representation in FER.

\subsection{Facial Expression Recognition}

As mentioned earlier, a canonical way of facial expression recognition (FER) is to extract facial expression representations, which are then mapped to confidence scores via a classifier \cite{dornaika2008simultaneous}. With this goal, numerous works \cite{zeng2018facial, li2021adaptively, she2021dive, lukov2022teaching, zheng2023attack} have achieved superior performance. For example, Wang \emph{et al.} \cite{wang2020region} proposed a region attention network to capture important areas for occlusion and pose variant FER. Zhao \emph{et al.} \cite{zhao2021learning} proposed a global multi-scale and local attention network for learning facial expression representations under occlusion and pose variation conditions. Xue \emph{et al.} \cite{Xue_2021_ICCV} explored Vision Transformer (ViT) \cite{dosovitskiy2021an} to learn diverse relation-aware local representations. Zeng \emph{et al.} \cite{Zeng_2022_CVPR} introduced unlabeled facial images to address the class imbalance problem in FER. Li \emph{et al.} \cite{li2022crs} designed a well-trained general encoder for learning facial expression representations, which can realize a linear evaluation on any target datasets. Zhao \emph{et al.} \cite{zhao2021robust} proposed a lightweight encoder for comprehensive facial representations. Wu \emph{et al.} \cite{Wu_2023_ICCV} leveraged facial landmarks to learn reliable facial expression representations with noisy labels. Zhang \emph{et al.} \cite{zhang2023leave} proposed an imbalanced FER method to extract transformation invariant information related to the minor categories from all training data. Nonetheless, most of existing FER methods learn representations using discrete labels, ignoring the emotional concept of different facial expressions. \emph{In this paper, we leverage text embeddings from the frozen VLM text encoder as an external knowledge for guiding facial expression representation learning.} 

\subsection{VLM in FER}

Recently, vision-language models (VLM) have demonstrated powerful potential in learning representations that bridge the visual and textual modalities through the joint training of two encoders. For example, Contrastive Language-Image Pre-training (CLIP) \cite{radford2021learning} consists of a visual encoder and a text encoder, which are trained with 400 million image-text pairs. It has demonstrated its excellent performance on learning visual representations in several downstream tasks \cite{menon2023visual, liu2023clip}. Motivated by this, several methods \cite{li2023cliper, zhao2023prompting, foteinopoulou2023emoclip} have explored learnable textual prompts instead of discrete labels for facial expression representation learning. For example, Li \emph{et al.} \cite{li2023cliper} proposed to learn a group of text descriptors for facial expression categories from the frozen CLIP. Zhao \emph{et al.} \cite{zhao2023prompting} fine-tuned the CLIP image encoder with learnable textual prompts for video-based FER. Tao \emph{et al.} \cite{tao20243} explored the alignment between the expression videos and abstract labels in the CLIP space. While the above methods are close in spirit to our work, they learn facial expression representations from the CLIP image encoder. \emph{To the best of our knowledge, our work is the first attempt to fine-tune an arbitrary pre-trained visual encoder using VLM text embeddings.}

\subsection{Disentangled Representation in FER}

Learning disentangled representations has been explored in FER for discriminative facial expression representations \cite{ruan2022adaptive}. For example, Yang \emph{et al.} \cite{dee_cvpr2018} claimed that a facial expression consists of an expressive component and a neutral face, and proposed to generate neutral faces with the generative adversarial network. Jiang \emph{et al.} \cite{jiang2022disentangling} regarded a facial representation as the combination of the identity, pose, and expression representations. Then, they combined the identity and pose representations for a neutral representation. Li \emph{et al.} \cite{li2023unconstrained} designed an encoder-decoder module to decompose a neutral face from a facial image. Similarly, Ruan \emph{et al.} \cite{Ruan_2021_CVPR} modeled facial expression information as the combination of the shared information across different categories and a unique information. Zhang \emph{et al.} \cite{Zhang_2021_CVPR} viewed the facial expression as the deviation from the identity. Indeed, existing methods mainly focus on learning neutral or shared information from facial images using discrete labels. \emph{In contrast, our method designs an emotional-to-neutral transformation via a text-guided process, which can further enhance facial expression representations.}

\section{Method}
\label{sec:method}

In this section, we first introduce the background of existing FER methods (Sec.~\ref{sec:background}). Then, we present the knowledge-enhanced FER pipeline (Sec.~\ref{sec:framework}). Finally, we describe the emotional-to-neutral transformation along with a self-contrast objective to further enhance facial expression representations (Sec.~\ref{sec:transformation}).

\begin{figure*}[t]
  \centering
   \includegraphics[width=1.0\linewidth]{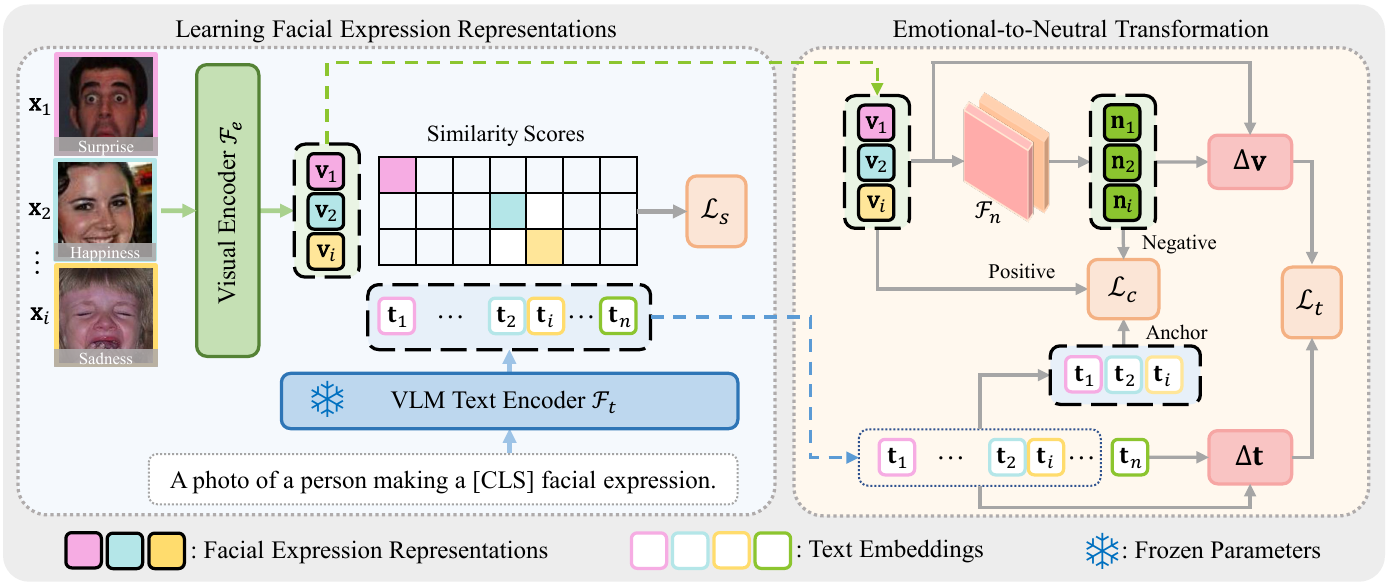}
   \caption{Illustration of the proposed method, whose core is to match facial expression representations from the visual encoder $\mathcal{F}_{e}$ with the corresponding text embeddings from the frozen VLM text encoder $\mathcal{F}_{t}$. Firstly, we calculate the similarity score between facial expression representation $\mathbf{v}_{i}$ and text embedding $\mathbf{t}_{i}$ via a cross-entropy loss $\mathcal{L}_{s}$. Then, we transform the facial expression representation $\mathbf{v}_{i}$ to a neutral representation $\mathbf{n}_{i}$ via a network $\mathcal{F}_{n}$. To achieve this, we measure the similarity between the representation difference $\Delta\mathbf{v}$ and the embedding difference $\Delta\mathbf{t}$ via a transformation loss $\mathcal{L}_{t}$. Finally, based on an anchor $\mathbf{t}_{i}$, a positive $\mathbf{v}_{i}$, and a negative $\mathbf{n}_{i}$, a self-contrast objective $\mathcal{L}_{c}$ constrains the distance between the text-expression representation pair $(\mathbf{t}_{i},\mathbf{v}_{i})$ and the text-neutral representation pair $(\mathbf{t}_{i},\mathbf{n}_{i})$. For clarity, we present three images from RAF-DB annotated with different categories.}
   \label{fig:2}
\end{figure*}

\subsection{Background}
\label{sec:background}

For a $C$-class FER task, there is a batch of training data $\mathcal{D}=\{(\mathbf{x}_{i}, y_{i})\}_{i=1}^{N}$, where $\mathbf{x}_{i}$ is the $i$-th training data, $y_{i}\in\{1,2,...,C\}$ is the corresponding label, and $N$ denotes the number of training data. Generally, the objective of FER is to fine-tune a pre-trained visual encoder for learning a facial expression representation $\mathbf{v}_{i}\in\mathbb{R}^{d_{v}\times1}$ by
\begin{equation}
  \mathbf{v}_{i}=\mathcal{F}_{e}(\mathbf{x}_{i};\theta_{e}),
  \label{eq:1}
\end{equation} 
where $\mathcal{F}_{e}$ denotes the visual encoder parameterized by $\theta_{e}$, and $d_{v}$ is the dimension of the representation. Meanwhile, a classifier is jointly trained to map the learned facial expression representation to confidence scores $\mathbf{p}_{i}\in\mathbb{R}^{C\times1}$ by
\begin{equation}
  \mathbf{p}_{i}=\mathcal{F}_{l}(\mathbf{v}_{i};\theta_{l}),
  \label{eq:2}
\end{equation} 
where $\mathcal{F}_{l}$ is the classifier parameterized by $\theta_{l}$. Finally, a cross-entropy loss is used to update both of these parameters with discrete labels.

As we discussed earlier, the discrete encoding via a classifier is a common training format in existing FER methods. However, the above supervision shares a major limitation in ignoring the emotional concept of different facial expressions. To address the limitation, we modify $\mathcal{F}_{l}$ to text embeddings from the frozen VLM text encoder, which contain rich knowledge.

\subsection{Knowledge-Enhanced FER}
\label{sec:framework}

The main objective of the FER task is to learn discriminative facial expression representations. We observe that text embeddings generated by the VLM text encoder can enhance to fine-tune a visual encoder for facial expression representations. In light of this observation, we propose a knowledge-enhanced method to formulate the FER problem as a process to match the similarity between the learned facial expression representation and VLM text embeddings. Our method can associate the discriminative information in facial expression representations with the rich knowledge in VLM text embeddings. 

As shown in Figure~\ref{fig:2}, the knowledge-enhanced method consists of two major steps. The first step extracts text embeddings from the frozen VLM text encoder. The second step leverages VLM text embeddings to fine-tune the visual encoder for facial expression representations. Specifically, given a prompt template (\emph{e.g.}, ``A photo of a person making a [CLS]\footnote{[CLS] is the category name, \emph{e.g.}, surprise, fear, disgust, happiness, sadness, anger, contempt, and neutral.} facial expression."), a text encoder can generate text embeddings $\mathbf{T}\in\mathbb{R}^{d_{t}\times C}$ by
\begin{equation}
  \mathbf{T}=\{\mathbf{t}_{c}=\mathcal{F}_{t}({\rm Prompt}_{c}), c=1,2,...,C\},
 \label{eq:text}
\end{equation}
where $\mathcal{F}_{t}$ is the VLM text encoder with frozen parameters, and ${\rm Prompt}_{c}$ denotes the prompt for the $c$-th category. We then normalize the facial expression representation $\mathbf{v}_{i}$ and the text embedding $\mathbf{t}_{c}$ for their similarity score as 
\begin{equation}
  {\rm sim}(\mathbf{t}_{c},\mathbf{v}_{i}) = \frac{\mathbf{t}_{c}
  \cdot\mathbf{v}_{i}}{||\mathbf{t}_{c}||||\mathbf{v}_{i}||}.
  \label{eq:3}
\end{equation}

Finally, we fine-tune the pre-trained visual encoder $\mathcal{F}_{e}$ via a cross-entropy loss:
\begin{equation}
  \mathcal{L}_{s} = -\frac{1}{N}\sum_{i=1}^{N}{\rm log}\frac{{\rm exp}({\rm sim}(\mathbf{t}_{i},\mathbf{v}_{i})/\tau)}{\sum_{c=1}^{C}{\rm exp}({\rm sim}(\mathbf{t}_{c},\mathbf{v}_{i})/\tau)},
  \label{eq:4}
\end{equation}
where $\mathbf{t}_{i}\in\mathbf{T}$ denotes the text embedding of the ground-truth category corresponding to the sample $\mathbf{x}_{i}$, namely the textual facial expression, and $\tau$ is a temperature parameter. As shown in Figure~\ref{fig:1}, the inter-class difference derived from the above process exhibits a great improvement compared to the process using discrete labels, but needs to be strengthened. Therefore, it is necessary to further enhance the discriminative power of facial expression representations.

\subsection{Emotional-to-Neutral Transformation}
\label{sec:transformation}

Recently, there is a psychological model (\emph{i.e.}, Russell’s Circumplex Model \cite{russell1980circumplex}) that different facial expressions fall on a two-dimensional circle, whose center precisely denotes the neutral category. In other words, we can leverage the center of the circle as a reference to better classify different facial expressions. Inspired by this, we propose an emotional-to-neutral transformation to derive a neutral representation from the facial expression representation itself. By a self-contrast objective among the neutral representation, the facial expression representation, and the textual facial expression, we can better fine-tune the pre-trained visual encoder. In the following, we will elaborate on these two components in details. 

\textbf{1) Transformation.} Unlike previous methods that either learn neutral representations using discrete labels \cite{dee_cvpr2018, jiang2022disentangling}, or generate neutral faces \cite{li2023unconstrained}, our method transforms a facial expression representation to a neutral representation by simulating the difference in text embeddings from textual facial expression to textual neutral. To achieve this, we introduce a network $\mathcal{F}_{n}$ parameterized by $\theta_{n}$ to transform a facial expression representation $\mathbf{v}_{i}$ to a neutral representation $\mathbf{n}_{i}\in\mathbb{R}^{d\times1}$ by 
\begin{equation}
  \mathbf{n}_{i}=\mathcal{F}_{n}(\mathbf{v}_{i};\theta_{n}).
  \label{eq:5}
\end{equation}

Then, we are inspired by the global direction \cite{dunlap2023using} to assume that the difference between the neutral category and facial expression is similar in both the representation space and text embedding space. Specifically, we define the difference $\Delta\mathbf{v}$ in the representation space and the difference $\Delta\mathbf{t}$ in the text embedding space as
\begin{equation}
  \Delta\mathbf{v}=\mathbf{v}_{i}-\mathbf{n}_{i},
  \label{eq:6}
\end{equation}
\begin{equation}
  \Delta\mathbf{t}=\mathbf{t}_{i}-\mathbf{t}_{n},
  \label{eq:7}
\end{equation}
where $\mathbf{t}_{n}$ denotes the text embedding corresponding to the neutral category, namely the textual neutral. Finally, we train $\mathcal{F}_{n}$ by encouraging the similarity between two normalized differences. Formally, a transformation loss is defined as
\begin{equation}
  \mathcal{L}_{t} = \frac{1}{N_{e}}\sum_{i=1}^{N_{e}}1-\frac{\Delta\mathbf{t}\cdot\Delta\mathbf{v}}{||\Delta\mathbf{t}||||\Delta\mathbf{v}||},
  \label{eq:8}
\end{equation}
where $N_{e}$ is the number of non-neutral faces in $\mathcal{D}$.

\textbf{2) Self-contrast.} The transformation loss in Eq.~(\ref{eq:8}) can derive a neutral representation from the learned facial expression representation. Therefore, we introduce a self-contrast objective to further enhance the discriminative power of facial expression representations. Specifically, we view the textual facial expression $\mathbf{t}_{i}$ as an anchor, the facial expression representation $\mathbf{v}_{i}$ as a positive, and the neutral representation $\mathbf{n}_{i}$ as a negative. Then, we maximize the similarity between the text-expression representation pair $(\mathbf{t}_{i}$,$\mathbf{v}_{i})$, and minimize the similarity between the text-neutral representation pair $(\mathbf{t}_{i},\mathbf{n}_{i})$. Formally, 
\begin{equation}
  \mathcal{L}_{c} = \frac{1}{N_{e}}\sum_{i=1}^{N_{e}}{\rm sim}(\mathbf{t}_{i},\mathbf{n}_{i})-{\rm sim}(\mathbf{t}_{i},\mathbf{v}_{i})+\gamma,
  \label{eq:9}
\end{equation}
where $\gamma$ is a parameter to ensure that the above loss is a non-negative value.

\textbf{3) Overall Objective Function.} The proposed method is optimized in an end-to-end process. The whole network parameterized by $\theta$ consisting of $\mathcal{F}_{e}$ and $\mathcal{F}_{n}$ minimizes the following loss function:
\begin{equation}
  \mathcal{L}_{total} = \lambda_{s}\mathcal{L}_{s}+\lambda_{t}\mathcal{L}_{t}+\lambda_{c}\mathcal{L}_{c},
  \label{eq:10}
\end{equation}
where $\lambda_{s}$, $\lambda_{t}$, and $\lambda_{c}$ are hyper-parameters to balance each term's intensity. The whole progress of our method is summarized in Algorithm~\ref{alg1}.

\begin{algorithm*}[t]
\caption{Main learning algorithm.}
\label{alg1}
\begin{algorithmic}[1]
   \Require Model parameters $\theta$, number of epochs $E_{max}$, number of iterations $I_{max}$, and learning rate $\eta$.
   \Ensure Updated model parameters $\theta$.
   \State \textcolor[RGB]{175,171,171}{// Training}
   \For{$E=1,2,3,...,E_{max}$}
   \For{$I=1,2,3,...,I_{max}$}
   \State Sample a training batch $\mathcal{D}=\{(\mathbf{x}_{i}, y_{i})\}_{i=1}^{N}$ randomly.
   \State Learning the facial expression representation $\mathbf{v}_{i}$ by Eq.~(\ref{eq:1}).
   \State Generate text embeddings $\mathbf{T}$ by Eq.~(\ref{eq:text}) and compute similarity scores by Eq.~(\ref{eq:3}).
   \State Compute $\mathcal{L}_{s}$ by Eq.~(\ref{eq:4}).
   \State \textcolor[RGB]{175,171,171}{// Emotional-to-neutral transformation}
   \If{$\mathbf{x}_{i}$ is not a neutral face}
   \State Obtain the neutral representation by Eq.~(\ref{eq:5}).
   \State Compute $\Delta\mathbf{v}$ and $\Delta\mathbf{t}$ by Eqs.~(\ref{eq:6}) and (\ref{eq:7}).
   \State Compute $\mathcal{L}_{t}$ and $\mathcal{L}_{c}$ by Eqs.~(\ref{eq:8}) and (\ref{eq:9}).
   \State Update $\theta \leftarrow \theta-\eta \nabla \mathcal{L}_{t}$.
   \State Update $\theta \leftarrow \theta-\eta \nabla \mathcal{L}_{c}$.
   \EndIf
   \State Update $\theta \leftarrow \theta-\eta \nabla \mathcal{L}_{s}$.
   \EndFor
   \EndFor
   \State \textcolor[RGB]{175,171,171}{// Testing}
   \State Deploy models $\mathcal{F}_{e}$ and $\mathcal{F}_{t}$ for the similarity score between representation $\mathbf{v}_{i}$ and embedding $\mathbf{t}_{c}$.
\end{algorithmic}
\end{algorithm*}

\section{Experiments}
\label{sec:experiments}

In this section, we conduct extensive experiments to verify the effectiveness of the proposed method. We first introduce the datasets (Sec.~\ref{sec:dataset}) and implementation details (Sec.~\ref{sec:implementation}). Then, we perform the ablation study to show the effect of each component in our method (Sec.~\ref{sec:ablation}). Finally, we compare the proposed method with state-of-the-art FER methods (Sec.~\ref{sec:sota}) and cross-dataset FER methods (Sec.~\ref{sec:cross}).

\subsection{Datasets}
\label{sec:dataset}

We evaluate the proposed method on four popular facial expression datasets, including RAF-DB, AffectNet, FERPlus, and CK+. 

\textbf{RAF-DB} \cite{Li_2017_CVPR} includes 29,672 real-world facial images, which are annotated by 40 annotators. In our experiments, we utilize facial images with six basic facial expression categories (\emph{i.e.}, surprise, fear, disgust, happiness, sadness, and anger) and a neutral category, consisting of 12,271 training data and 3,068 testing data. The overall accuracy and the mean accuracy across all categories are reported by default. 

\textbf{AffectNet} \cite{mollahosseini2017affectnet} is currently the largest facial expression dataset, containing about 450,000 facial images, which are manually annotated with 11 categories. We conduct experiments on seven categories and eight categories. Specifically, for the 7-class (7 cls), there are 283,901 training images and 3,500 validation images (500 images per class). For the 8-class (8 cls), there are 287,568 training images and 4,000 validation images (500 images per class). Since it suffers from a significant imbalance, we follow the setting \cite{wang2020suppressing} to utilize the same oversampling strategy for a fair comparison. The overall accuracy is reported by default.

\textbf{FERPlus} \cite{BarsoumICMI2016} is a large-scale facial expression dataset collected via Google image search APIs. It provides annotations for seven facial expression categories (\emph{i.e.}, six basic categories and the contempt) and the neutral category. All images are annotated by 10 crowd-sourced taggers, consisting of 28,709 training images, 3,589 validation images, and 3,589 testing images. The overall accuracy on testing data is reported by default. 

\textbf{CK+} \cite{lucey2010extended} contains 593 video sequences from 123 subjects. For a fair comparison, we select the first and last frame of each sequence as the neutral face and the targeted facial expression, respectively. It consists of 636 facial images annotated with six basic categories and the neutral category. The overall accuracy is reported by default.

\subsection{Implementation Details}
\label{sec:implementation}

The proposed method is implemented based on PyTorch with one NVIDIA V100 GPU. To take a fair comparison with results, we use ResNet-18 \cite{he2016deep} and Tiny Swin Transformer (Swin-T) \cite{liu2021swin} both pre-trained on MS-Celeb-1M dataset \cite{guo2016ms} as the visual encoder. The frozen VLM text encoder is used to generate text embeddings for different facial expression categories and the neutral category. 

During training, we use MTCNN \cite{zhang2016joint} to align and resize facial images to 224$\times$224 pixels for ResNet-18 and 112$\times$112 pixels for Swin-T. Each model is fine-tuned with the SGD optimizer for 50 epochs. The initial learning rate of 2e-3 is adjusted with a warm-up cosine scheduler. The batch size, momentum, and weight decay are set as 64, 0.9, and 5e-4, respectively. Following a standard setting \cite{Xue_2021_ICCV}, an data augmentation strategy, including random rotate and crop, random horizontal flip, and random erasing, is applied in all experiments. In Eq.~(\ref{eq:10}), the hyper-parameters $\lambda_{s}$, $\lambda_{t}$, and $\lambda_{c}$ are set to 1.0, 1.0, and 1.0, respectively. We set the default $\gamma$ as 2. The temperature parameter $\tau$ is set to 0.01. The transformation network $\mathcal{F}_{n}$ is a 2-layer MLP with input and output dimensions of 512 and a hidden dimension of 128. Considering the changing dimension values of text embeddings from different VLM, we adjust the dimension $d_{v}$ to be equal to the dimension $d_{t}$ using a fully-connected layer. During testing, we employ the visual encoder and the frozen VLM text encoder for similarity scores between facial expression representations and text embeddings.

\subsection{Ablation Study}
\label{sec:ablation}

In this section, we analyze the effect of different VLM text encoders, different pre-trained visual encoders, different loss funtions in our method, different prompt templates, and varying balancing hyper-parameters. 

\begin{table*}[t]
  \centering
  \caption{Ablation study of text embeddings generated from different VLM text encoders for fine-tuning ResNet-18 and Swin-T via $\mathcal{L}_{s}$ on RAF-DB, AffectNet (7 cls and 8 cls), and FERPlus. ``Acc'' and ``mean Acc'' denote the overall accuracy (in \%) and the mean accuracy (in \%). This also applies to the following tables.}
  \begin{tabular}{c|c|cc|cc|c}
    \toprule[1pt]
    \multirow{2}{*}{Encoder} & \multirow{2}{*}{VLM}& \multicolumn{2}{c|}{RAF-DB} & \multicolumn{2}{c|}{AffectNet} & \multicolumn{1}{c}{FERPlus}\\\cmidrule{3-4} \cmidrule{5-6} \cmidrule{7-7} ~ & ~&Acc&mean Acc&Acc (7 cls)&Acc (8 cls)&Acc\\
    \hline
    \multirow{3}{*}{ResNet-18}
    & ALIGN \cite{jia2021scaling} & 85.72 & 75.37 & 62.43 & 59.83 & 87.06 \\
    & BLIP \cite{li2022blip} & 86.21 & 76.06 & 62.34 & 59.80 & 87.57 \\
    & CLIP \cite{radford2021learning} & 87.26 & 80.94 & 63.71 & 60.23 & 88.33\\
    \hline
    \multirow{3}{*}{Swin-T}
    & ALIGN \cite{jia2021scaling} & 89.47 & 80.97 & 63.91 & 60.48 & 87.63 \\
    & BLIP \cite{li2022blip} & 89.31 & 80.83 & 64.00 & 60.53 & 87.82 \\
    & CLIP \cite{radford2021learning} & 90.55 & 82.43 & 64.71 & 61.50 & 88.93 \\
    \bottomrule[1pt]
    \end{tabular}
  \label{tab:VLM}
\end{table*}

\textbf{Effect of Different VLM Text Encoders.} In this work, we leverage VLM text embeddings as the external knowledge to guide facial expression representation learning. To investigate the impact of text embeddings generated from different VLM text encoders, we conduct an ablation study using three popular VLMs, including CLIP \cite{radford2021learning}, ALIGN \cite{jia2021scaling}, and BLIP \cite{li2022blip}. As shown in Table~\ref{tab:VLM}, we observe that CLIP \cite{radford2021learning} consistently achieves superior performance. Therefore, we use the frozen CLIP as the default VLM for generating text embeddings in the following experiments. 

\textbf{Effect of Different Pre-trained Visual Encoders.} To further verify the effectiveness of fine-tuning a visual encoder using text embeddings, Table~\ref{tab:loss} compares three types of pre-trained visual encoders, including the ResNet-18 pre-trained on MS-Celeb-1M (rows 1 to 5), the best available ViT-B/16 in CLIP (rows 6 to 10), and the Swin-T pre-trained on MS-Celeb-1M (rows 11 to 15). From this table, it clearly shows that text embeddings are more powerful compared to the classifier-based FER pipeline using discrete labels. For example, compared with the classifier-based results (rows 1, 6, and 11), the results supervised by text embeddings via $\mathcal{L}_{s}$ (rows 2, 7, and 12) can surpass them by 0.62\%, 1.43\%, and 0.88\% on RAF-DB, respectively.

Besides, CLIP \cite{radford2021learning} is pre-trained with 400 million curated image-text pairs containing a wide variety of visual concepts, but risks learning face-independent information \cite{li2023clip}. Compared with the ViT-B/16-based results, the Swin-T-based results outperform them in each case. The remarkable results demonstrate that fine-tuning the visual encoder with parameters pre-trained on a large-scale face dataset is more helpful in learning discriminative facial expression representations.  

\begin{table}[t]
  \centering
  \caption{Ablation study of three loss functions on RAF-DB and AffectNet (8 cls) using different pre-trained visual encoders, including ResNet-18, ViT-B/16 in CLIP, and Swin-T. Rows 1, 6, and 11 denote that a classifier and the encoder are jointly trained using discrete labels via the cross-entropy loss.}
  \begin{tabular}{c|ccc|c|c}
    \toprule[1pt]
    Encoder & $\mathcal{L}_{s}$ & $\mathcal{L}_{t}$ & $\mathcal{L}_{c}$ & RAF-DB & AffectNet\\
    \hline
    \multirow{5}{*}{ResNet-18}
    &-&-&-& 86.64 & 59.48  \\
    &\checkmark&-&-& 87.26 & 60.23 \\
    &\checkmark&\checkmark&-& 88.53 & 60.60  \\
    &\checkmark&-&\checkmark& 88.82 & 60.85  \\
    &\checkmark&\checkmark&\checkmark& 89.86 & 61.25 \\
    \hline
    \multirow{5}{*}{ViT-B/16}
    &-&-&-& 88.59 & 59.83\\
    &\checkmark&-&-& 90.02 &  60.58 \\
    &\checkmark&\checkmark&-& 90.45 & 61.68  \\
    &\checkmark&-&\checkmark& 91.04 & 61.95  \\
    &\checkmark&\checkmark&\checkmark& 91.36 & 62.18  \\
    \hline
    \multirow{5}{*}{Swin-T}
    &-&-&-&89.67& 60.25 \\
    &\checkmark&-&-& 90.55 & 61.50  \\
    &\checkmark&\checkmark&-& 91.33 & 62.88  \\
    &\checkmark&-&\checkmark& 91.62 & 62.92  \\
    &\checkmark&\checkmark&\checkmark& 92.63 & 63.90  \\
    \bottomrule[1pt]
    \end{tabular}
  \label{tab:loss}
\end{table}

\begin{figure}[t]
  \centering
    \includegraphics[width=0.75\linewidth]{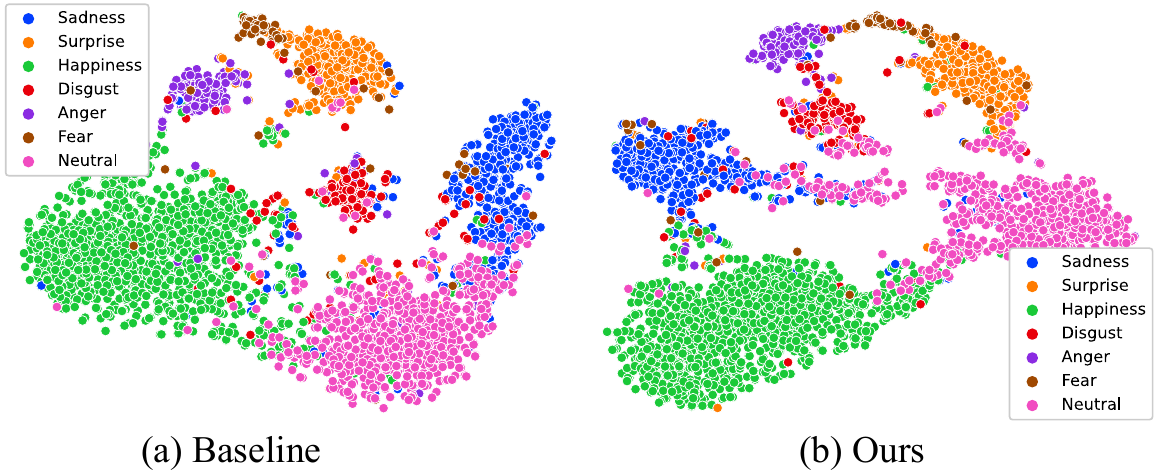}
   \caption{2D t-SNE visualization  \cite{maaten2008visualizing} of facial expression representations extracted from the RAF-DB testing set using ResNet-18 in different manners, including fine-tuning via (a) $\mathcal{L}_{s}$ and (b) the combination of $\mathcal{L}_{s}$, $\mathcal{L}_{t}$, and $\mathcal{L}_{c}$.}
   \label{fig:3}
\end{figure}

\textbf{Effect of Different Loss Functions.} To better understand the role of each loss function in our method, we carry out the ablation study of the gradual addition of different loss functions into the baseline using three visual encoders on RAF-DB and AffectNet (8 cls). As shown in Table~\ref{tab:loss}, several observations can be summarized as follows: 1) Compared with the Swin-T-based baseline via $\mathcal{L}_{s}$ (row 12), transforming the facial expression representation to a neutral representation (row 13) slightly improves the performance by 0.78\% and 1.38\%. Another addition of $\mathcal{L}_{c}$ improves the performance to 91.62\% and 62.92\%; 2) A significant improvement of 2.08\% and 2.40\% is achieved after the addition of combining $\mathcal{L}_{t}$ and $\mathcal{L}_{c}$. These results validate the contribution of the emotional-to-neutral transformation along with the self-contrast objective for learning discriminative facial expression representations. 

In addition to the above quantitative analysis, we present t-SNE visualization results about the distribution of facial expression representations. As shown in Figure~\ref{fig:3}, our method can achieve a clear tendency to push facial expression categories away from the neutral category. This observation demonstrates the effectiveness of the self-contrast objective, which can further enhance the discriminative power of facial expression representations.

\begin{figure}[t]
  \centering
   \includegraphics[width=0.8\linewidth]{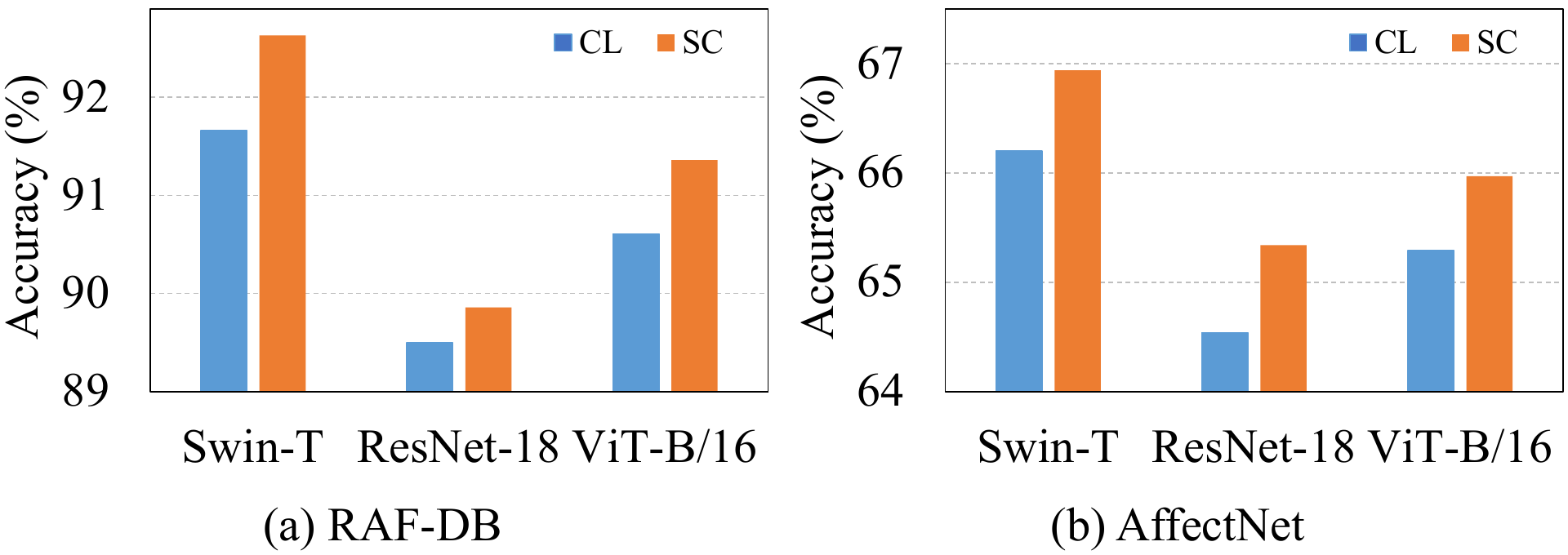}
   \caption{Evaluation of different functions $\mathcal{L}_{c}$, including contrastive learning (CL) and self-contrast (SC) objective using the pre-trained Swin-T, ResNet-18, and ViT-B/16 in CLIP on (a) RAF-DB and (b) AffectNet (7 cls).}
   \label{fig:4}
\end{figure}

\textbf{Effect of Different Functions $\mathcal{L}_{c}$.} In this work, we introduce a self-contrast objective to constrain the relationship among the anchor (textual facial expression), the positive (facial expression representation), and the negative (neutral representation). Similarly, contrastive learning \cite{chen2020simple} is a loss function that can achieve the above goal. Figure~\ref{fig:4} compares two loss functions on RAF-DB and AffectNet (7 cls). We can observe that the self-contrast objective outperforms the contrastive learning in each case. For example, compared with the contrastive learning, the self-contrast objective on RAF-DB using the pre-trained Swin-T obtains a larger margin by 0.97\%. This suggests that the self-contrast objective contributes to fine-tuning the visual encoder. 

\begin{table*}
  \centering
  \caption{Evaluation of different prompt templates using the pre-trained ResNet-18 and Swin-T on RAF-DB, AffectNet (7 cls), and FERPlus (in \%, overall accuracy).}
  \begin{tabular}{c|c|ccc}
    \toprule[1pt]
    Encoder&Prompt Templates & RAF-DB & AffectNet &FERPlus \\
    \hline
    \multirow{9}{*}{ResNet-18}
    & This person is [CLS]. & 89.08 & 64.80 & 88.27\\
    & A photo of a [CLS] face. & 89.28 & 64.51 & 89.15\\
    & This is a [CLS] facial expression. & 89.21 & 64.49 & 89.23\\
    & A person makes a [CLS] facial expression. & 88.72 & 64.51 & 89.53\\
    & An ID photo of a [CLS] facial expression. & 88.59 & 64.54 & 89.63\\
    & A person is feeling [CLS]. & 89.05 & 64.57 & 89.09\\
    & A person feeling [CLS] on the face. & 88.53 & 64.37 & 89.28\\
    & A photo of a person with a [CLS] expression on the face. & 88.72 & 64.77 & 89.41\\
    & A photo of a person making a [CLS] facial expression. & 89.86 & 65.34 & 90.36\\
    \hline
    \multirow{9}{*}{Swin-T}
    &This person is [CLS]. & 91.92 & 65.31 & 89.34\\
    &A photo of a [CLS] face. & 91.63 & 65.17 & 90.23\\
    &This is a [CLS] facial expression. & 91.46 & 65.34 & 89.79\\
    &A person makes a [CLS] facial expression. & 91.82 & 65.37 & 90.17\\
    &An ID photo of a [CLS] facial expression. & 91.75 & 65.77 & 89.72\\
    &A person is feeling [CLS]. & 91.43 & 66.00 & 89.79\\
    &A person feeling [CLS] on the face. & 91.23 & 65.86 & 89.95\\
    &A photo of a person with a [CLS] expression on the face. & 91.66 & 65.89 & 89.42\\
    &A photo of a person making a [CLS] facial expression. & 92.63 & 66.94 & 91.18\\
    \bottomrule[1pt]
    \end{tabular}
  \label{tab:template}
\end{table*}

\textbf{Effect of Different Prompt Templates.} As aforementioned that text embeddings significantly outperform discrete labels for learning facial expression representations, we mark that the choice of prompt templates is critical. In this work, we design several prompt templates based on OpenAI's report \cite{goh2021multimodal} and the experience. Table~\ref{tab:template} shows the effects of nine types of prompt templates using different visual encoders. We can observe that the default template of ``A photo of a person making a [CLS] facial expression." consistently achieves the best performance in each case.

\begin{figure}[t]
  \centering
   \includegraphics[width=0.8\linewidth]{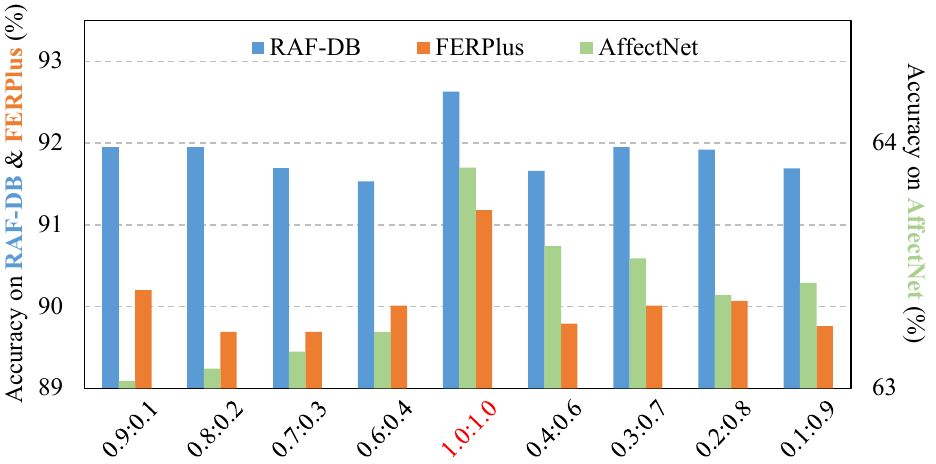}
   \caption{Evaluation of different forms of balancing hyper-parameters ($\lambda_{t}$:$\lambda_{c}$) using the pre-trained Swin-T on RAF-DB, AffectNet (8 cls), and FERPlus. The performance with the default setting is marked in the red.}
   \label{fig:hyper}
\end{figure}

\textbf{Effect of Varying Balancing Hyper-parameters.} 
In Eq.~(\ref{eq:10}), there are three hyper-parameters $\lambda_{s}$, $\lambda_{t}$, and $\lambda_{c}$ to balance three loss functions. To further examine the effect of three hyper-parameters, we conduct experiments with varying balancing parameters. Note that we default $\lambda_{s}=1.0$ as the baseline with the cross-entropy loss $\mathcal{L}_{s}$ for a fair comparison. Figure~\ref{fig:hyper} shows the performance comparison of different $\lambda_{t}$ and $\lambda_{c}$. Obviously, the result demonstrates that the default setting achieves the best performance compared to other settings.

\subsection{Comparison with State-of-the-Art Methods}
\label{sec:sota}

\begin{table*}
  \centering
  \caption{Performance comparison with state-of-the-art FER methods using the pre-trained ResNet and ViT on RAF-DB, AffectNet (7 cls and 8 cls), and FERPlus. ``Text'' denotes learning facial expression representations using text embeddings, otherwise using discrete labels via a classifier. Baseline denotes that two pre-trained visual encoders (\emph{i.e.}, ResNet-18 and Swin-T) are fine-tuned using text embeddings via $\mathcal{L}_{s}$. This also applies to the following tables.}
  \begin{tabular}{c|c|c|c|cc|cc|c}
    \toprule[1pt]
    \multirow{2}{*}{Encoder} & \multirow{2}{*}{Method }& \multirow{2}{*}{Params}& \multirow{2}{*}{Text} & \multicolumn{2}{c|}{RAF-DB} & \multicolumn{2}{c|}{AffectNet} & \multicolumn{1}{c}{FERPlus}\\\cmidrule{5-6} \cmidrule{7-8} \cmidrule{9-9} ~ &~ & ~ & ~&Acc&mean Acc&Acc (7 cls)&Acc (8 cls)&Acc\\
    \hline
    \multirow{9}{*}{ResNet}
    & SCN \cite{wang2020suppressing} & 11.2M &$\huge$$\times$& 88.14 & 76.40 & - & 60.23 & 88.01\\
    & RUL \cite{zhang2021relative} & 11.2M&$\huge$$\times$& 88.98 & 81.66 & 61.56 & 55.08 & 88.30\\
    & MA-Net \cite{zhao2021learning} & 11.2M&$\huge$$\times$& 88.40 & 81.29 & 64.53 & 60.29 & 88.71\\
    & DACL \cite{Farzaneh_2021_WACV} & 11.2M & $\huge$$\times$& 87.78 & 80.44 & 65.20 & - & 88.20\\
    & Face2Exp \cite{Zeng_2022_CVPR} & 23.5M &$\huge$$\times$& 88.54 & - & 64.23 & - & -\\
    & EAC \cite{zhang2022learn} & 11.2M &$\huge$$\times$& 89.50 & 81.84 & 65.32 & 60.53 & 89.64\\
    %& MTAC \cite{liu2023uncertain} (TMM 2023) &$\huge$$\times$& 89.31 & - & - & 61.58 & 88.74\\
    %& GAAVE \cite{zheng2023attack} (AAAI 2023) && 89.29 &  81.69 & 65.60 & 62.78 & 89.83 \\
    & Latent-OFER \cite{Lee_2023_ICCV} & 11.2M &$\huge$$\times$& 89.60 & - & 63.90 & - & -\\\cmidrule{2-9}
    & Baseline & 11.3M &$\checkmark$& 87.26 & 80.94 & 63.71 & 60.23 & 88.33\\
    & Ours & 11.3M &$\checkmark$& \textbf{89.86} & \textbf{82.11} & \textbf{65.34} & \textbf{61.25} & \textbf{90.36}\\
    \hline
    \multirow{9}{*}{ViT}
    & VTFF \cite{ma2021facial} & 51.8M &$\huge$$\times$& 88.14 & 81.20 & 64.80 & 61.85 & 88.81\\
    & MVT \cite{li2021mvt} & 60.3M &$\huge$$\times$& 88.62 & 80.38 & 64.57 & 61.40 & 89.22\\
    & TransFER \cite{Xue_2021_ICCV} & 65.2M &$\huge$$\times$& 90.91 & 85.86 & 66.23 & - & 90.83\\
    & AU-ViT \cite{mao2022aware} & 89.3M &$\huge$$\times$ &91.10 & 84.57&65.59&-&90.15\\
    & APViT \cite{xue2022vision} & 65.2M &$\huge$$\times$& 91.98 & 86.36 & 66.86 & - & 90.86 \\
    & FER-former \cite{li2023fer} & - &$\checkmark$& 91.30 & 85.43 & - & - & 90.96 \\ 
    & CLIPER \cite{li2023cliper} & 86.3M &$\checkmark$& 91.61 & - & 66.29 & 61.98 & -\\\cmidrule{2-9}
    %& POSTER \cite{Zheng_2023_ICCV} (ICCV 2023) &$\huge$$\times$& 92.05 & 86.03 & \textbf{67.31} &  63.34 & 91.62\\\cline{2-8}
    & Baseline & 28.6M &$\checkmark$& 90.55 & 82.43 & 64.71 & 61.50 & 88.93 \\
    & Ours & 28.6M &$\checkmark$& \textbf{92.63} & \textbf{87.06} & \textbf{66.94} & \textbf{63.90} & \textbf{91.18}\\
    \bottomrule[1pt]
    \end{tabular}
  \label{tab:sota}
\end{table*}

To validate the effectiveness of our method, we conduct experiments on RAF-DB, AffectNet, and FERPlus to compare with state-of-the-art methods in two aspects, including ResNet-based and ViT-based visual encoders. As shown in Table~\ref{tab:sota}, several observations can be summarized as follows: 1) Our method outperforms existing methods both using the pre-trained ResNet and ViT encoder. For example, compared with APViT \cite{xue2022vision}, our method achieves the improvement of 0.65\% (overall accuracy) and 0.70\% (mean accuracy) on RAF-DB. Note that existing methods mainly learn facial expression representations using discrete labels. Differently, our method can enhance facial expression representations using text embeddings; 2) To the best of our knowledge, CLIPER \cite{li2023cliper} could be the first facial expression recognition method using CLIP. Compared with it, the proposed method improves the overall accuracy by over 1.02\% on RAF-DB, 0.65\% and 1.92\% on AffectNet, respectively. These results demonstrate the effectiveness of our method in learning facial expression representations; 3) We also compare the number of encoder parameters between our method and existing FER methods. From the results, except for Face2Exp \cite{Zeng_2022_CVPR} using the pre-trained ResNet-50, we use the same pre-trained ResNet-18 as most of existing methods. Note that the increasing 0.1M parameters originate from the transformation network $\mathcal{F}_{n}$. Besides, compared with ViT-based methods, our method achieves superior performance on different datasets with fewer parameters.

\begin{figure*}[t]
  \centering
   \includegraphics[width=1.0\linewidth]{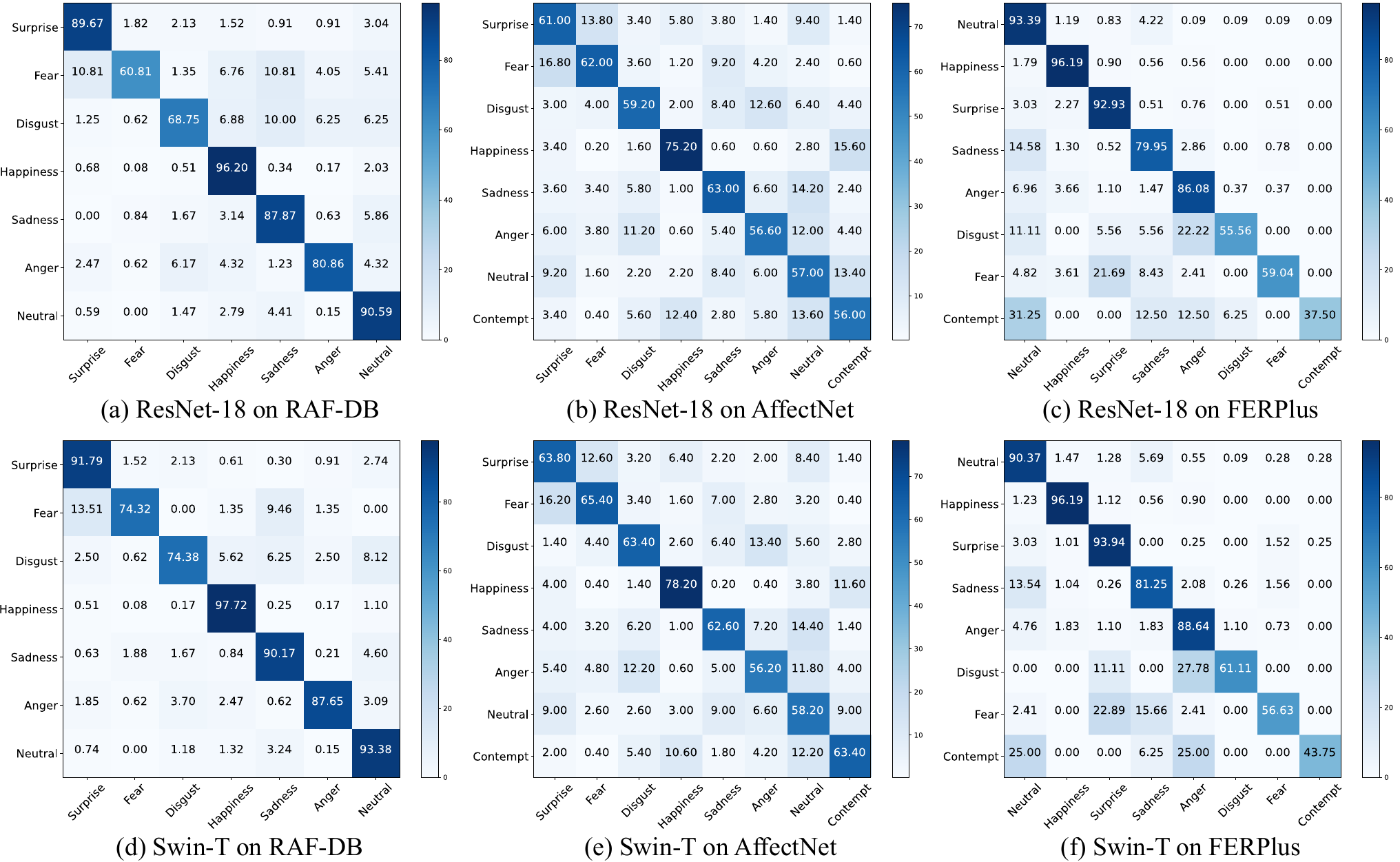}
   \caption{The confusion matrices of our method using the pre-trained ResNet-18 and Swin-T on RAF-DB, AffectNet, and FERPlus datasets.}
   \label{fig:confusion}
\end{figure*}

\begin{table}[t]
  \centering
  \caption{Cross-dataset evaluation using the pre-trained ResNet and ViT on the lab-collected CK+ dataset (in \%, overall accuracy). All models are trained on RAF-DB and tested on the CK+ dataset.}
  \begin{tabular}{c|c|c|c}
    \toprule[1pt]
    Encoder & Method &Text& CK+\\
    \hline
    \multirow{8}{*}{ResNet}
    & gACNN \cite{li2019occlusion} (2019) &$\huge$$\times$& 81.07 \\
    & AGRA \cite{chen2021cross} (2022) &$\huge$$\times$& 77.52 \\
    & Ada-CM \cite{li2022adacm} (2022) &$\huge$$\times$& 85.32 \\
    %& CSRL \cite{chen2023cross} (2023) &&  88.37\\
    & SPWFA-SE \cite{spwfa2023} (2023) &$\huge$$\times$& 81.72 \\
    & DENet \cite{li2023unconstrained} (2023) &$\huge$$\times$&82.55\\
    & TAN \cite{ma2023transformer} (2023) &$\huge$$\times$& 82.64 \\\cmidrule{2-4}
    & Baseline &$\checkmark$& 85.85 \\
    & Ours &$\checkmark$& \textbf{87.89} \\
    \hline
    \multirow{5}{*}{ViT}
    & VTFF \cite{ma2021facial} (2023) &$\huge$$\times$& 81.88 \\
    & APViT \cite{xue2022vision} (2022) &$\huge$$\times$& 86.64 \\
    & POSTER \cite{Zheng_2023_ICCV} (2023) &$\huge$$\times$& 85.53 \\\cmidrule{2-4}
    & Baseline &$\checkmark$& 86.12 \\
    & Ours &$\checkmark$& \textbf{88.36} \\
    \bottomrule[1pt]
    \end{tabular}
  \label{tab:cross}
\end{table}

In addition, we visualize the confusion matrices of our method using the pre-trained ResNet-18 and Swin-T on RAF-DB, AffectNet, and FERPlus. As shown in Figure~\ref{fig:confusion}, our method achieves satisfactory performance on most of facial expression categories, especially on Happiness. We also observe the relatively poor performance on Fear, Disgust, and Contempt. This might be explained by the reason that most of existing facial expression datasets are imbalanced \cite{zhang2023leave}. For example, the training data of Fear and Happiness in RAF-DB consists of 281 and 4,772 facial images, respectively. 

\subsection{Cross-dataset Evaluation}
\label{sec:cross}

Text embeddings serve as a way to describe shared information across different facial expression datasets. To verify the generalization ability of our method, we present an evaluation from training on RAF-DB to testing on CK+, which is a widely-used scheme for cross-dataset FER. Table~\ref{tab:cross} compares the quantitative results between our method and existing methods using the pre-trained ResNet and ViT. Our method demonstrates a significant performance improvement compared to other methods that utilize discrete encoding to learn facial expression representations. Specifically, compared with the previous best results, our method achieves an improvement of 2.57\% using the pre-trained ResNet-18, and 1.72\% using the pre-trained ViT, respectively. It suggests that our method learns facial expressions representations using text embeddings, which are general for different facial expression datasets.

\begin{figure}[t]
  \centering
   \includegraphics[width=0.75\linewidth]{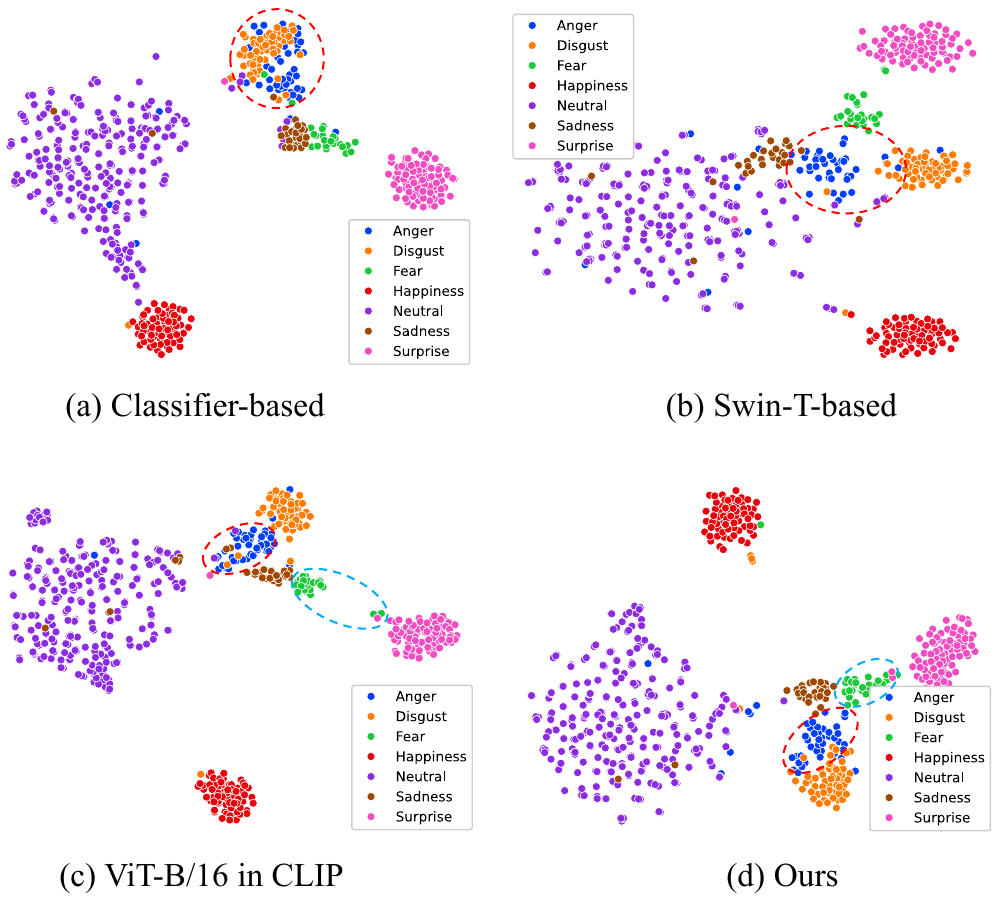}
   \caption{2D t-SNE visualization of facial expression representations extracted from the CK+ dataset by different methods, including (a) the classifier-based result with Swin-T, (b) the Swin-T-based result via $\mathcal{L}_{s}$, and fine-tuning (c) ViT-B/16 in CLIP and (d) Swin-T via the combination of $\mathcal{L}_{s}$, $\mathcal{L}_{t}$, and $\mathcal{L}_{c}$.}
   \label{fig:5}
\end{figure}

In addition, we conduct visualization experiments to further evaluate the proposed method. Figure~\ref{fig:5} shows the distribution of facial expression representations using different methods on the CK+ dataset. We can observe that the representations extracted by two baseline strategies are not easily distinguishable for some facial expression categories, \emph{e.g.}, the anger in the red dotted line. In contrast, our method effectively enhances the inter-class difference and the intra-class similarity. Especially, compared with ViT-B/16 in CLIP, our method achieves a more distinct intra-class compactness for the fear representations in the blue dotted line. 

\section{Conclusion and Future Work}
\label{sec:conclusion}

While fine-tuning visual encoders pre-trained on a face dataset using discrete labels becomes prevalent for the FER task, leveraging VLM text embeddings to fine-tune the visual encoder has not been explored. In this paper, a new knowledge-enhanced FER method is proposed to match the similarity between a facial expression representation and VLM text embeddings. Meanwhile, we propose an emotional-to-neutral transformation to derive a neutral representation from the facial expression representation itself via a text-guided process. Together with the transformation, we further introduce a self-contrast objective to enhance the discriminative power of facial expression representations. Extensive experiments on four popular facial expression datasets demonstrate the effectiveness of our method using VLM text embeddings as the supervision signal. 

In the future, we will continue to focus on how to transform dynamic facial expression representations to the neutral representation using VLM text embeddings. The core problem might be how to distinguish irrelevant frames in videos, which may belong to other facial expression categories. Besides, the proposed method limits in generating VLM text embeddings with manually-designed prompt templates. We will discuss the effect of the learnable textual prompt \cite{zhou2022learning}. 

%Bibliography
\bibliographystyle{unsrt}  
\bibliography{references}

\end{document}